\documentclass[sigconf]{aamas} 

\usepackage{balance} 
\usepackage{soul}
\usepackage{algorithmicx}
\usepackage[linesnumbered,ruled,vlined]{algorithm2e}
\usepackage{tabularx}
    \newcolumntype{L}{>{\raggedright\arraybackslash}X}
\SetKwInOut{Parameter}{Parameters}
\SetKw{KwBy}{by}

\setcopyright{ifaamas}
\acmConference[AAMAS '23]{Proc.\@ of the 22nd International Conference
on Autonomous Agents and Multiagent Systems (AAMAS 2023)}{May 29 -- June 2, 2023}
{London, United Kingdom}{A.~Ricci, W.~Yeoh, N.~Agmon, B.~An (eds.)}
\copyrightyear{2023}
\acmYear{2023}
\acmDOI{}
\acmPrice{}
\acmISBN{}

\acmSubmissionID{???}

\title{ShelfHelp: Empowering Humans to Perform Vision-Independent Manipulation Tasks with a Socially Assistive Robotic Cane}

\author{Shivendra Agrawal}
\affiliation{
  \institution{University of Colorado Boulder}
  \city{Boulder}
  \state{Colorado}
  \country{USA}}
\email{shivendra.agrawal@colorado.edu}

\author{Suresh Nayak}
\affiliation{
  \institution{University of Colorado Boulder}
  \city{Boulder}
  \state{Colorado}
  \country{USA}}
\email{suresh.nayak@colorado.edu}

\author{Ashutosh Naik}
\affiliation{
  \institution{University of Colorado Boulder}
  \city{Boulder}
  \state{Colorado}
  \country{USA}}
\email{ashutosh.naik@colorado.edu}

\author{Bradley Hayes}
\affiliation{
  \institution{University of Colorado Boulder}
  \city{Boulder}
  \state{Colorado}
  \country{USA}}
\email{bradley.hayes@colorado.edu}

\begin{abstract}
The ability to shop independently, especially in grocery stores, is important for maintaining a high quality of life. This can be particularly challenging for people with visual impairments (PVI). Stores carry thousands of products, with approximately 30,000 new products introduced each year in the US market alone, presenting a challenge even for modern computer vision solutions. Through this work, we present a proof-of-concept socially assistive robotic system we call ShelfHelp, and propose novel technical solutions for enhancing instrumented canes traditionally meant for navigation tasks with additional capability within the domain of shopping. 
ShelfHelp includes a novel visual product locator algorithm designed for use in grocery stores and a novel planner that autonomously issues verbal manipulation guidance commands to guide the user during product retrieval. Through a human subjects study, we show the system's success in locating and providing effective manipulation guidance to retrieve desired products with novice users. We compare two autonomous verbal guidance modes achieving comparable performance to a human assistance baseline and present encouraging findings that validate our system's efficiency and effectiveness and through positive subjective metrics including competence, intelligence, and ease of use. 
\end{abstract}


\newcommand{\BibTeX}{\rm B\kern-.05em{\sc i\kern-.025em b}\kern-.08em\TeX}

\begin{document}


\pagestyle{fancy}
\fancyhead{}


\maketitle 


\section{Introduction}
\begin{figure}[tp]
    \centering
    \includegraphics[width=242pt]{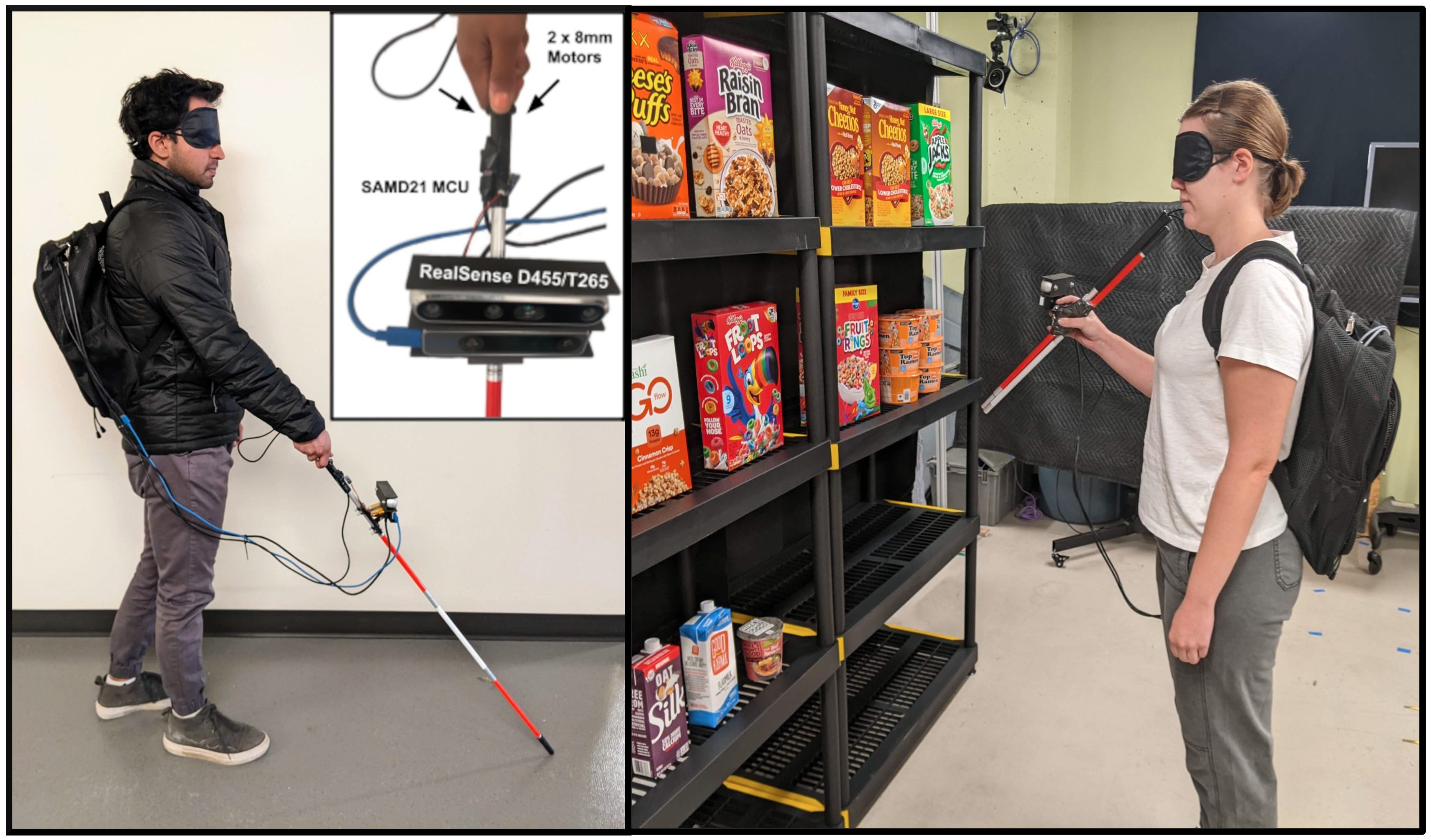}
    \caption{ShelfHelp includes a robotic cane equipped with RealSense D455 and T265 cameras. The system is powered through a laptop in a backpack. \textit{Left:} The system used as a navigational device. It uses audio and haptic feedback for navigation guidance. \textit{Right:} The system used as a manipulation device. It uses audio for manipulation guidance.
    }
    \label{fig:device}
\end{figure}

It is estimated that 295 million people have some form of vision impairment, of whom 43.3 million are blind \cite{bourne2021trends}. Currently, when encountering difficulty with shopping, they can get help from sighted humans. This lack of independence has other negative connotations such as loss of privacy \cite{ahmed2015privacy}.
Some PVI shoppers have also indicated they are not willing to use store staffers for shopping for items that require discretion such as medicine and personal hygiene items \cite{ahmed2015privacy,kulyukin2005robocart,kulyukin2006ergonomics}. Our work seeks to alleviate the dependence on guide availability and to mitigate the loss of privacy encountered with traditional support mechanisms.
People have also expressed the need to have a system that can help them locate items in a store and even at their home \cite{ahmed2015privacy}. Moreover, some environments present additional challenges while solely relying on tactile sensing. For example - 1) Kitchen counters may contain hot or sharp objects and may pose a safety issue. 2) Grocery stores with similar items shelved together or with an otherwise dense concentration of items contributes to poor tactile differentiability.

Grocery shopping primarily consists of three main subtasks: navigation, product retrieval, and product examination. This work focuses on product retrieval. Prior research on spatial cognition of PVI uses a dichotomous ontology where they distinguish the space around the person in two categories: \textit{1) locomotor} and \textit{2) haptic} \cite{golledge1996cognitive}. Locomotor space is the space around the user whose exploration and access requires locomotion whereas haptic space is the immediate space around the user that can be sensed without bodily translation. Our work addresses the research problem in the haptic space of the user.

We propose two fine-grain manipulation guidance systems that use verbal instructions to guide the user towards retrieving the desired item. We use the grocery store domain because of the representative challenges it contains and the opportunity for immediate positive broader impact. We also introduce a novel product detection algorithm to locate desired items in a grocery setting with a hand-held camera system. In this work, we showcase the manipulation guidance system as part of a grocery shopping assistant. We define fine-grain manipulation guidance with respect to object retrieval as guidance that brings the user close to the desired object such that the object is in the haptic space of the user and is accessible to the user's grasp. Our solution aims to minimize the exhaustive search of the object in the haptic space. This is especially inefficient and a burden for PVI in a grocery store because of the high density of products that could be present even in the limited haptic space of the person. 

The hardware choices for ShelfHelp are motivated by utility, cost, and extensibility. Our system extends the capability of a robotic smart cane \cite{agrawal2022novel} created originally for social navigation assistance (Fig.\ref{fig:device}). This area has been extensively researched \cite{saha2019closing, Real-surveyOfNavigation, Takizawa-Kinect, Chen-CCNYSmartCane, zhang_pose_estimation, Saaid-rangeRecognition, Niitsu-dangerousObstacles, Murali-visuallyChallenged, Wang-independentNavigation, Sakhardande-ultrasonic1, Wahab-audioandvibe, Megalingam-voiceandvibe, singh_kapoor_2020, Meshram-Objectdetection, varghese-wheels, Nasser-Thermalcane, Ulrich-wheels, chung-smartPhone, Guerrero-CaBot, xiao-GuideDog,Katzschmann-vibeVest,slade2021multimodal, wachaja2017navigating}, yet many important capabilities are still rich for exploration. It is practical and prudent to utilize the sensing and compute power of these existing devices to address multitude of critical tasks for a more independent lifestyle.

To summarize, this paper contributes,
\begin{itemize}
    \item A robotic cane system that leverages computer vision to assist in independent grocery shopping for PVI.
    \item A modular two stage computer vision pipeline to locate desired products in a grocery setting.
    \item A novel fine-grain manipulation guidance system that optimizes for guide time and the number of commands without compromising legibility.
    \item A pilot study validating the system's success in locating and providing effective guidance to retrieve a desired object from a dense cluster of objects with novice users.
    \item Findings on user preferences regarding desirable planner properties.
\end{itemize}

\section{Related Work}
\subsection{Manipulation guidance}
Manipulation guidance is an area that has been explored within the robotics community for over a decade. Vasquez et al.\cite{vazquez2014assisted} showed that saliency maps could be used to find regions of interest (ROI) and directed users' hand to the ROI. They found that their verbal commands' efficacy suffered because they did not utilize a global frame of reference. Bonani et al. \cite{bonani2018my} showed promise for the concept with an experimenter-controlled teleoperated system and Bigham et al. \cite{bigham2010vizwiz} did so with a Mechanical Turk-based system, but fully autonomous implementations were outside the scope of their contributions. Their manipulation guidance guided people to the general direction of the product but the system didn't focus on fine-grain manipulation guidance which is important in many scenarios, for example a kitchen countertop or where tactile differentiability makes exhaustive search inefficient such as grocery stores which have similar items densely situated. The most popular solution in this problem domain is a human-powered service called Be My Eyes \cite{bemyeyes}, but this service suffers from scalability issues due to its reliance on human availability, is not readily available in developing countries, requires an active data connection, and introduces nigh-unavoidable privacy concerns. We present a novel verbal guidance solution wherein we learn a mapping of language commands to human hand movements and map them to actions within a Markov Decision Process (MDP) that can be solved with well-established reinforcement learning techniques, informing our guidance of the user.

\subsection{Product identification}
Existing techniques with a fixed number of output classes work with a limited database of products in a grocery store, for example, Feng et al. \cite{feng2020research} trained a system for classifying 1329 products. These techniques may be impractical for the scale required of a solution for this domain because of the sheer amount of products \cite{NielsenIQ} available and the slight variations they come in. It is infeasible to require creation and maintenance of labeled data and the training of an object classifier. Existing technology that is reliable in product identification uses bar code scanning \cite{gharpure2008robot, lookout}. This often needs internet connectivity and can only identify a product once it is in the person's possession, making it inefficient for search over the grocery shelves. Our work is inspired by the body of work in \emph{instance retrieval}, where the task is to find a target image in the scene \cite{chen2021deep}. Our proposed product identification system does not require re-training and works offline, making it scalable and practical.

\subsection{Grocery assistant systems}
Recent work on grocery assistant systems focuses largely on navigation inside the store \cite{gharpure2008robot, kulyukin2010accessible,kulyukin2005robocart,kulyukin2006ergonomics,nicholson2009shoptalk}, also known as the locomotor space of the user. Solutions that rely on environmental augmentation, such as the addition of RFID tags and barcodes, introduce barriers to adoption as they are inapplicable in uninstrumented domains. Barcodes alone can not reliably be used to locate a product from a dense cluster of products on a grocery shelf, as the shopper can typically only scan a barcode once they have retrieved the product. Prior research \cite{gharpure2008robot} has also focused on text-based product selection as an input method. Researchers have also focused on other issues around shopping, including identifying products that users are running out of and organizing newly purchased products at home \cite{yuan2019constructing}, as well as ``the last few meters'' way-finding problem \cite{saha2019closing} and solutions for people with low vision \cite{zhao2016cuesee}. 
We focus on an unsolved research area that primarily considers the haptic space of the user.  

To summarize, our solution addresses 1) the problem of locating a desired product and 2) the challenge of providing effective fine-grain verbal guidance to reach and grasp the product.

\section{System Design}
ShelfHelp extends the capabilities of an existing robotic cane originally developed as a navigational assistance device. This was a design choice to re-purpose an existing hardware system since many of the navigational assistance prototypes have the necessary sensing and compute (Figure \ref{fig:device}). The system is capable of locating desired products and verbally guiding the user to retrieve them. The software system follows a serial architecture composed of three main components - \textit{perception, planning, } and \textit{guidance}.

\subsection{Hardware System}\label{hardware}
The hardware (Fig \ref{fig:device}) consists of a cane-mounted RealSense D455 for RGB-Depth sensing and RealSense T265 for odometry. The sensors are connected to a Dell G15 laptop with an RTX 3060 GPU carried in a backpack worn by the user. As a navigational assistance device, the system is held with a standard cane grip but as a manipulation assistance device, a different, less collision prone grasp is used behind the sensors.

\subsection{Software System}\label{software}
\begin{figure}[h]
    \centering
    \includegraphics[width=240pt]{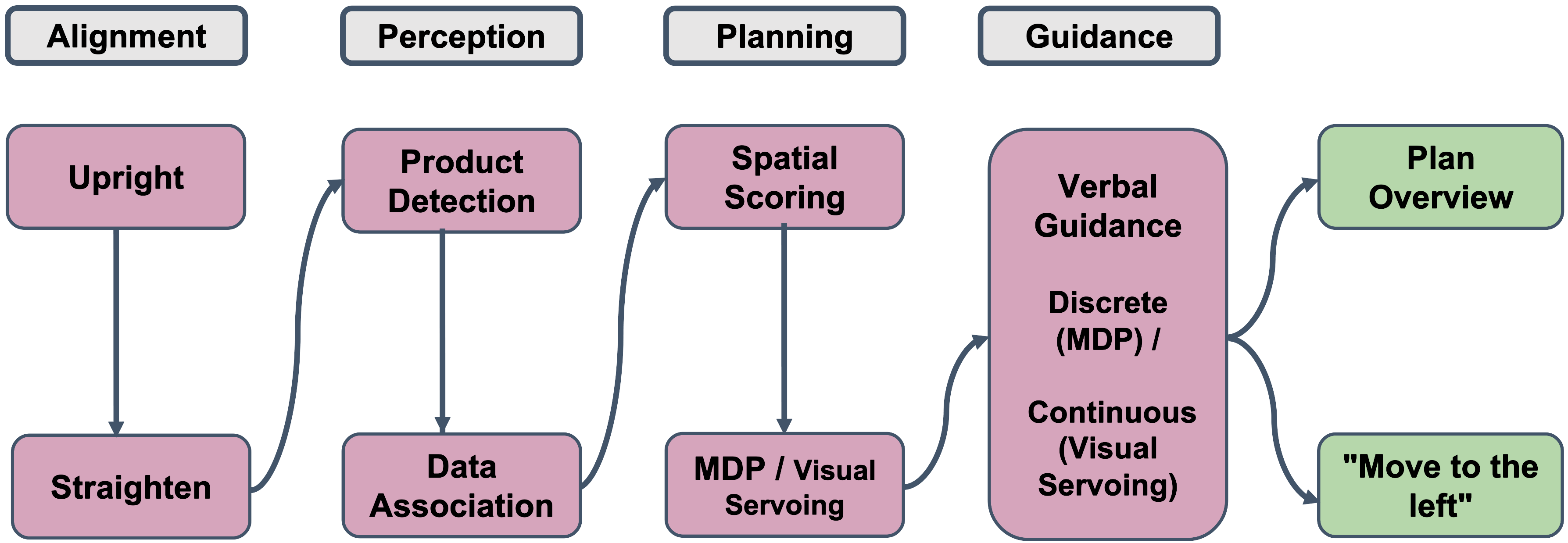}
    \caption{System Diagram. Alignment, perception, planning, and verbal conveyance are executed on a backpack-worn laptop, while all the sensing is mounted on the cane.}
    \label{fig:system}
\end{figure}

Our work can best be partitioned into sections regarding innovations in perception, solutions to the data association problem for maintaining a consistent and persistent mapping of product detections over time, product selection, fine-grain manipulation planning to reach the selected product, and methods for conveying this manipulation plan to the human user to complete the task.

\subsubsection{Alignment}\label{alignment}
In order for the system to issue commands in the user's egocentric frame of reference, our system issues verbal commands to align the user and the system with respect to the shelf. We use the T265 camera's IMU to guide the user to point the system straight at the shelf in an upright way. We also align the user so that they directly oppose the shelf normal. This is done by locating the shelf after camera upright alignment via Hough transform. The system issues commands to turn in place until the largest horizontal line in the Hough transform is almost completely horizontal in the camera frame. 

\subsubsection{Product Detection}\label{perception}
\begin{algorithm}
\DontPrintSemicolon 

\KwIn{scene, target\_product, camera\_pose}
\KwOut{best\_product}
\Parameter{\texttt{similarity\_threshold}}
$boundingbox\_list \gets region\_proposal(scene)$\;
$target\_feature \gets feature\_extractor(target\_product)$\;
$similar\_instances \gets []$\;
\ForEach{$bb \in boundingbox\_list$}{
    $bb\_feature \gets feature\_extractor(bb)$\;
    $bb\_score \gets similarity(target\_feature, bb\_feature)$\;
    /* \textit{Scene has RGB and depth information}\;
    $bb\_pose \gets camera\_to\_world(bb, scene, camera\_pose)$\;
    \uIf{$bb\_score \ge \texttt{similarity\_threshold}$} {
        $similar\_instances.append(bb\_pose, bb\_score)$
    }
}
$similar\_products \gets data\_association(similar\_instances)$\;
$best\_product \gets spatial\_scoring(similar\_products)$\;
\Return{\text{best\_product}}\;
\caption{{\sc Product Detection Procedure}}
\label{algo:detect}
\end{algorithm}

\begin{figure}[h]
    \centering
    \includegraphics[width=245pt]{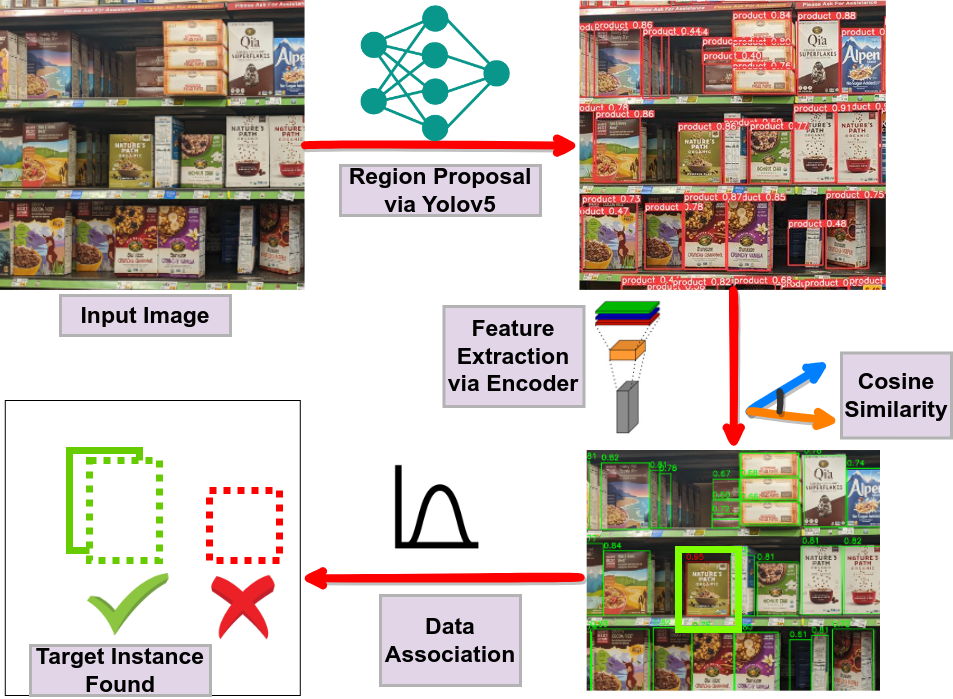}
    \caption{Our product search algorithm can reliably locate desired products on a grocery shelf. Regions with a high likelihood of containing any product are proposed in the first stage. The features of these regions are then compared against the target product image. Our data association solution is used to identify whether detections from incoming camera frames are new or re-detections of existing products. The above image shows our algorithm operating within an actual grocery store, where the product classification aspect of this work has been tested and validated. The data association and manipulation assistance components were validated within a lab-based study.
    }
    \label{fig:result}
\end{figure}

The entire product detection pipeline works in realtime making it appropriate for real world usage. To use this system, we require that the user has only a single image of the product that they want to find. This image can be acquired in a number of ways, for example by taking a picture the first time it is purchased or by downloading an image from the internet. We have developed a novel two-stage product search system (Algorithm \ref{algo:detect}). 
\begin{itemize}
    \item In the first stage, our method proposes regions in form of bounding boxes that are most likely to contain \emph{any} product (line 1 of Algorithm \ref{algo:detect}). We train the YoloV5 network on the SKU-110K dataset  \cite{goldman2019dense} to create a product detector. This network is robust and generates region proposals with 0.91-precision and 0.77-recall upon cross-validation with the SKU-110K dataset. Figure \ref{fig:result} (upper-right) shows a sample result from a real grocery store. 
    
    \item In the second stage, an encoder is used as a feature extractor (line 2 and 5) that matches the features of the proposed regions and the target image, finding the best possible match (Fig \ref{fig:result}). We do this by training an autoencoder on MS-COCO color images \cite{lin2014microsoft} and then utilizing the encoder portion as the feature extractor (Fig \ref{fig:autoencoder}). This method doesn't require any retraining and works in real-time. For each new image added to the database, we pass it through the frozen encoder and save the generated feature vector on disk. This avoids requiring any retraining of the autoencoder. The encoder finds the latent space representation of the target product image and each of the proposed regions.  We compare the representation vectors using  \textit{cosine similarity} to find closer vectors (line 6). We empirically determine a similarity score threshold (0.6) that captures satisfactory performance across real-world environments, but this value can easily be fine-tuned in case there is a significant distribution shift between the evaluation and deployment environments. 
    \begin{figure}[h]
    \centering
    \includegraphics[width=230pt]{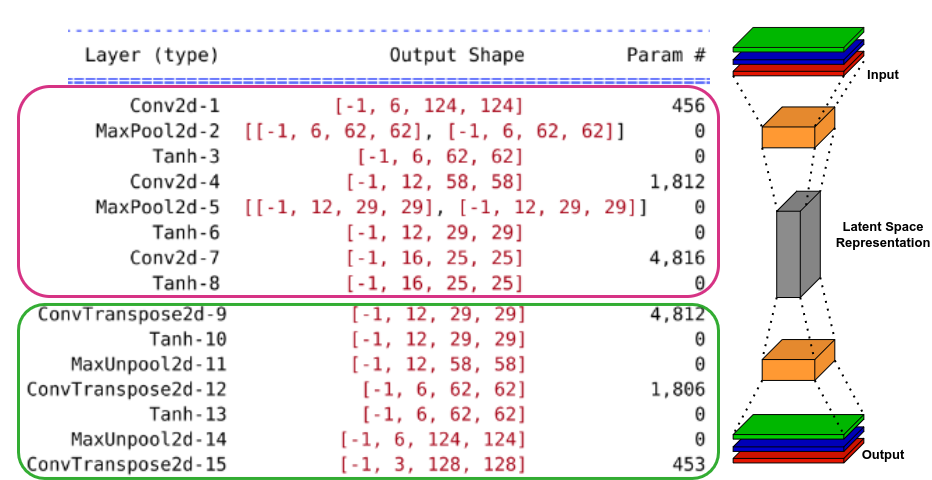}
    \caption{The architecture of the \textit{autoencoder} trained for feature extraction. ShelfHelp uses the \textit{encoder} portion (top layers outlined in red) as the feature extractor that creates a latent space representation of the images.}
    \label{fig:autoencoder}
    \end{figure}
\end{itemize}
To transform the information from the camera frame to a fixed global frame, we use pose information obtained by a cane-mounted RealSense T265 (which has minimal drift in indoor settings) running an onboard Simultaneous Localization and Mapping (SLAM) algorithm, and fuse it with the depth information obtained from a cane-mounted RealSense D455 (line 8). We use a Gaussian Mixture Model (GMM) to refine detections and distinguish between the foreground and background depth information, as the bounding boxes can contain significant background pixels in case a product and its bounding box are not overlapping significantly.

\subsubsection{Data Association}\label{dataassociation}
We use data association techniques to identify each instance of the same product uniquely across subsequent frames of camera capture (line 11). This is particularly challenging because similar products exist in groups. We do this by defining each product instance as a multivariate Gaussian defined by the tuple $p = \{x_g, y_g, z_g, w, h \}$ where $x_g, y_g, z_g$ is the 3D location of the product in a global frame and $w, h$ are width and height in meters. Accounting for the width and the height helps us discard some incorrect matches, as incorrectly proposed regions with our method not only have to have similar features but also similar shape to be incorrectly labeled (Fig. \ref{fig:result} - lower left). The IMU data from the T265 sensor helps to calculate the object's pose in a global frame of reference and helps to create a persistent ``map" of the product location. This way we can align the verbal directional commands with respect to the current hand pose and avoid the drawback of formulating verbal commands generated with targets located only in the camera frame of view, which is sensitive to hand movements \cite{vazquez2014assisted}. Product instance information is updated using a rolling mean over associated detections. We also employ a \textit{lazy deletion strategy} to delete instances that have not been seen a sufficient number of times (sparse detections) or recently (old detections).

\subsubsection{Spatial Scoring}\label{scoring}
\begin{figure}[h]
    \centering
    \includegraphics[width=200pt]{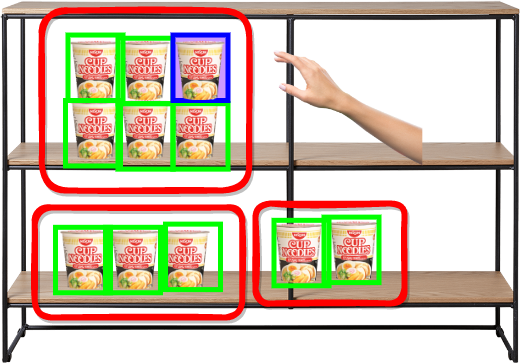}
    \caption{The spatial scoring system clusters all the found instances spatially 
 and gives preference to the closest cluster to the current hand pose. Ties are broken arbitrarily.}
    \label{fig:cluster}
\end{figure}

ShelfHelp then scores each detected product instance of the target product and picks the one with the highest score as its planning goal (line 12). It considers the rolling similarity with the target image and the spatial information. This allows the system to utilize important information that is absent without an explicit physics model, namely selecting an instance from the top level of stacked items to minimize the risk of toppling (Fig \ref{fig:cluster}). 

\subsubsection{Planning}\label{planning}
We have developed two different guidance mechanisms to provide verbal instruction once the target has been located: \textit{1) continuous}, for example, \textit{``keep on going right"...``stop''} and \textit{2) discrete}, for example, \textit{``move 6 inches to the right''}. 
The decision to create a discrete guidance mode was inspired by study results showing that PVI have been known to perceive length units better than sighted people \cite{andreou2005estimation} and the criteria of minimizing verbal feedback for the task as research has shown that PVI prefer not being provided with excessive information \cite{skovfoged2022there}.

\subsubsection{Continuous Planner}\label{continuous}
The continuous guidance operates by calculating the relative position of the target and the device (Fig. \ref{fig:device}), providing continuous cues along each individual axis of movement until the next is aligned. It generates the commands with the following template: \textit{``keep on going \{direction\}"} where direction can be \textit{\{left, right, up, down, forward, backward\}}. The system issues the command \textit{``keep on going"} if it notices the hand slow down sufficiently in anticipation of the next command. Once the error on the current guidance axis is lower than a threshold the system issues the command \textit{``stop''}. 

\subsubsection{Dataset from Human Demonstrations:}
\begin{figure}[t]
    \centering
    \includegraphics[width=230pt]{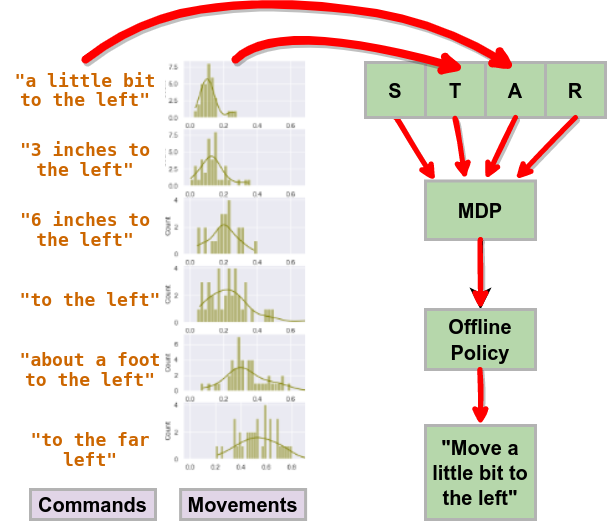}
    \caption{\textit{(Left to right)} A sample of discrete commands. The movement (in meters) each command caused. MDP and solution definition. We train a model of human hand movement from demonstrations that inform the transition probabilities $T$. $S$ defines the state space, $A$ defines the discrete set of verbal actions, and $R$ is the reward function. A policy is learned offline that can be used across reaching tasks.
    }
    \label{fig:displacement}
\end{figure}

To develop the discrete guidance method, we collected a dataset mapping verbal movement commands from a fixed command-set, recording participants' net hand movements upon reacting to that command (Fig \ref{fig:displacement}). We formed $36$ discrete commands and issued $1250$ instances of the commands in total to $25$ volunteers ($50$ commands per person) while they were blindfolded. The hand movement data were recorded using an OptiTrack motion capture system. Figure \ref{fig:displacement} shows a sample from this dataset illustrating commands pertaining to \textit{left} movement. We can see that the hand movement caused by these commands is close to a Gaussian distribution and the \textit{mean} increases for the verbal commands that convey a larger movement. We fitted Gaussians to characterize the movement caused by each command as 
\[
X_c \sim \mathcal{N}(\mu_c,\,\sigma_{c}^{2})\,
\]
where $\mu_c$ and $\sigma_c$ are the mean and standard deviation of the movement caused by command $c$.


\subsubsection{Discrete Planner}\label{discrete}
Through the human-subjects demonstrations, we find that the hand movement following the commands was not deterministic. Moreover, a greedy approach would not guarantee an optimal solution, akin to the 0-1 Knapsack problem. We solve this sequential decision-making problem by modeling it as an MDP formulated with $(S,A,T,R)$ where $S$ is the set of states in the MDP, $A$ is the set of actions, $T$ is a stochastic transition function describing the action-based state transition dynamics of the model, and $R$ is a reward function.
\begin{itemize}
\item $S$ is the tuple $({\Delta}x, {\Delta}y, {\Delta}z, axis)$ where the first three terms are the difference in distance of the target and the hand pose, and \emph{axis} defines the axis of motion for the previous command which could be any of three values corresponding to the X (horizontal), Y (vertical), or Z (depth) axis and a direction \textit{\{left, right, up, down, forward, backward\}}. This formulation is dependent on the relative distance between the target and the hand and thus it finds a plan for all the potential states that can be encountered. We discretized the states at 5 cm resolution and considered a cuboid region of 0.8m as the operational space for the human hand. The system assumes the hand to be at the boundary of our cuboid in case it was outside of the region. In practice, a region bigger than this produced the same policy for regions that were further away.
\item $A$ is the set of discrete verbal commands such as \textit{``Move 6 inches to the left"}. 
\item $T$ is calculated from $X_c$ as the movement caused by each command is not deterministic. We use the Euclidean difference between the states and Gaussian $X_c$ associated with the command $c$ to find out the probability of the transition between those states when command $c$ is issued. 
\item The reward function $R$ encourages reaching the target and discourages issuing superfluous commands. It also discourages a sequence of commands that could be illegible or frustrating by penalizing axis changes. 

Table \ref{table:reward} shows the reward values and their description.
The $axis\_order\_reward$ rewards adherence to a prescribed ordering of guidance with respect to the axes. For example, we set the initial guidance axis as \textit{vertical} and the planner rewards transitioning from the \textit{vertical} axis to the \textit{horizontal} axis and not the \textit{depth} axis. This helps to hone in on the product on the XY plane and then reach out for the product in the \textit{depth} axis. The $living\_penalty$ penalizes excessive number of commands. The $interleaving\_penalty$ penalizes excessive interleaving of axes that could make the guidance illegible. It is necessary and suitable to transition once the distance to the target on the current axis of guidance is reduced and short enough. We ensure this by setting $necessary\_transition\_reward$ inversely proportional to the distance remaining (error) on the current guidance axis.


\begin{table}[h]
\centering
    \caption{Outline of MDP Reward values \label{table:reward}}
    \begin{tabularx}{\linewidth}{|p{40mm}|L|} 
    \hline
        \textbf{Reward} & \textbf{Description} \\ 
    \hline
        $goal\_state\_reward = +10000$ & reward for reaching the goal \\
    \hline
        $living\_penalty = -10$ & penalty for using a command \\
    \hline
        $interleave\_penalty = -100$ & penalty for changing axis of command \\
    \hline
        $axis\_order\_reward = +100$ & reward for transitioning from vertical to horizontal axis \\
    \hline
        $necessary\_transition\_reward =$ \newline $ \left(0.001+\text{current axis error}\right)^{-1}$ & reward for axis transition when current axis error is removed \\
    \hline
    \end{tabularx}
\end{table}
\end{itemize}

The MDP is then solved using value iteration to generate a general reaching policy that can be queried online to guide the user toward arbitrary target locations. 

\subsubsection{Guidance}\label{guidance}
The verbal guidance has two components. It begins with a plan overview that is followed by hand movement instructions.

\textit{1) Plan Overview:} The actual guidance is preceded by a plan overview which serves to set the expectations for the general direction of the forthcoming instructions. We do so by computing the initial heading direction to the target. We project and discretize the heading direction to the target onto the plane of the shelf and use clock-based direction. The overview has the following template: \textit{``I found the product at about \{ \} o'clock direction"}.

\textit{2) Hand Movement Instructions:}
The plan overview is followed by guidance instructions. The instructions to convey to the user (actions) are computed online when using the continuous guidance mode and queried online from the policy learned in the discrete guidance mode. Based on the relative position of the target and the hand, a command is formulated (or retrieved from the policy) and issued aloud from a speaker that is part of the robotic cane system. In an effort to reduce frustration, the system issues new commands only when the user's hand has slowed down sufficiently to show that they are ready for the next command. The user is asked to grasp the target object with their non-occupied hand if they are close to it with the system.

\section{Pilot Study}

We conducted an IRB-approved study with novice users (n=15) for system testing and validation. Participants were all sighted students who were blindfolded for the testing. At this stage of design and technical readiness, we follow prior work in performing preliminary validations using blindfolded users \cite{slade2021multimodal,wachaja2017navigating,dosSantos-comparativeStudy,o2014detachable,fukasawa2012navigation,okazaki2014perceived} as a precursor to engaging with the PVI community.

\subsection{Hypotheses}
We test the following hypotheses in our comparison of the two planner modes. 
\begin{itemize}
	\item \textbf{H1}: Participants will retrieve products with a lesser number of commands in the discrete guidance mode compared to the continuous guidance mode.
	\item \textbf{H2}: Participants will retrieve the products in less time with the discrete guidance mode as compared to the continuous guidance mode, once the products are located.
	\item \textbf{H3}: Participants will retrieve the products with less net hand movement with the discrete guidance mode compared to the continuous guidance mode. 
\end{itemize}

\begin{figure}[t]
    \centering
    \includegraphics[width=200pt]{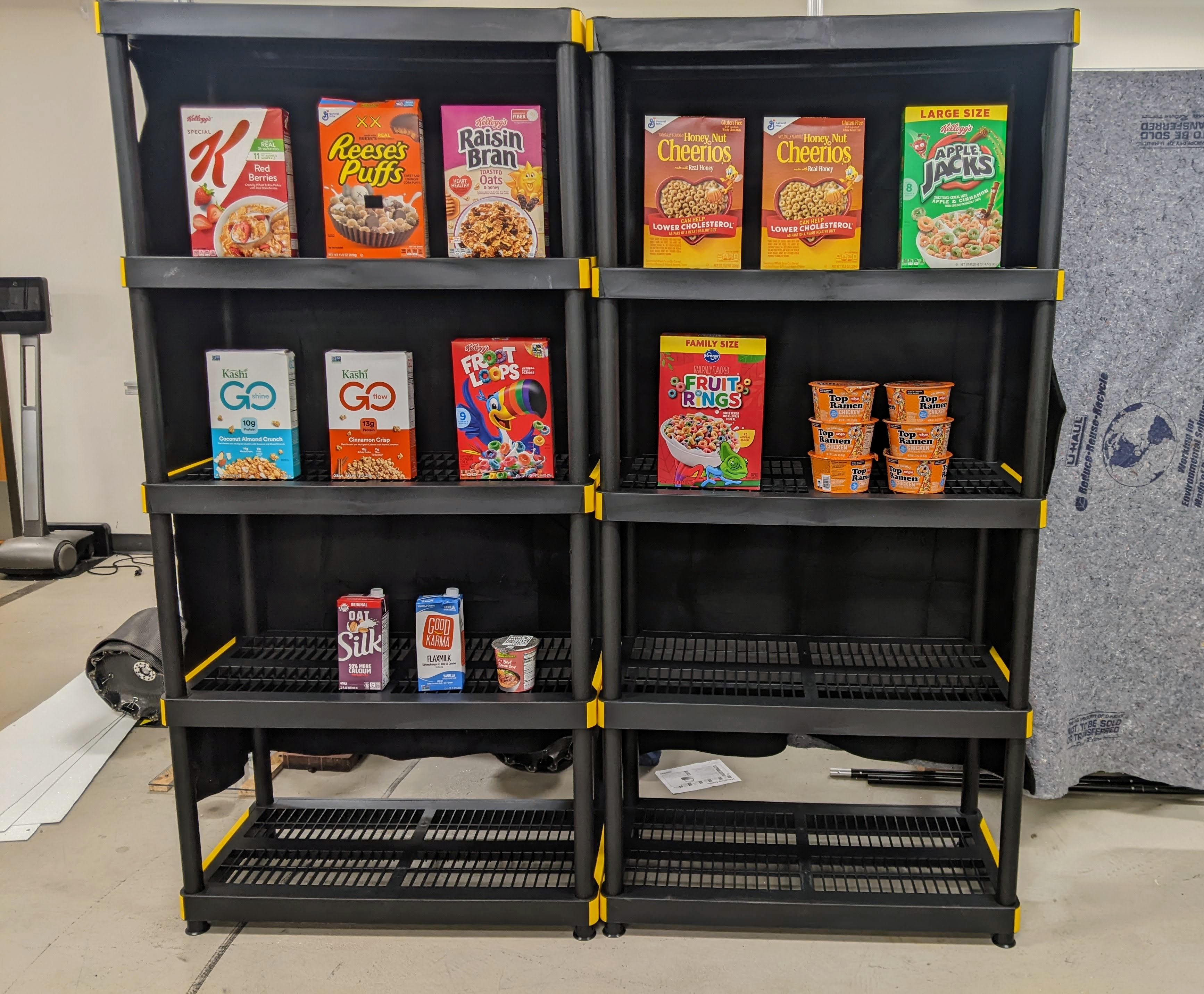}
    \caption{The experimental setup approximating a grocery store shelf, used for evaluating the efficacy of our manipulation guidance system.}
    \label{fig:setup}
    \vspace{-4mm}
\end{figure}

\subsection{Experiment Design}
We use the experimental setup shown in Figure \ref{fig:setup} to test the system. While we test the system as a whole, we do so with a specific focus on the two proposed guidance planners - 1) Continuous and 2) Discrete. As a baseline, we also compare our system against a human caller (university research staff separate from the experimenters) guiding over a video call. The caller gave freeform verbal guidance with the aim of helping the participant reach the goal on the shelf. The users performed each of these 3 conditions (continuous, discrete, and human) 5 times. We had a set of 5 different products for each of the conditions and the same set of 5 products with the same spatial configuration was used for each condition. The products were picked from the top two rows. We randomly shuffled the 3 conditions, counterbalancing as needed to achieve parity of orderings to avoid training/familiarity biases. We ensured that the users have the shelf in the frame at the starting pose and they were approximately 0.9m to 1.5 away from the target product in all the runs of the experiment. Each user was oriented on how to use the system, how to hold it, and how many runs they would perform. We asked them to keep their locomotion to a minimum and primarily move their hand. We also informed them about the alignment process and how to interpret the alignment guidance instructions that precede the actual guidance. The orientation process took about 3-4 minutes on average for each user.

\subsection{Results}
\subsubsection{System Success Rate}
\begin{figure}[h]
    \centering
    \includegraphics[width=150pt]{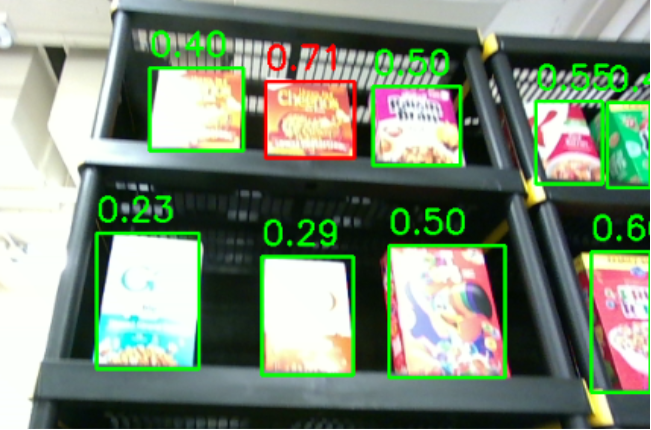}
    \caption{A sample product detection result that was robust to blurring and overexposure with the help of \textit{data association}.}
    \label{fig:blur}
\end{figure}

\begin{figure}[h]
    \centering
    \includegraphics[width=200pt]{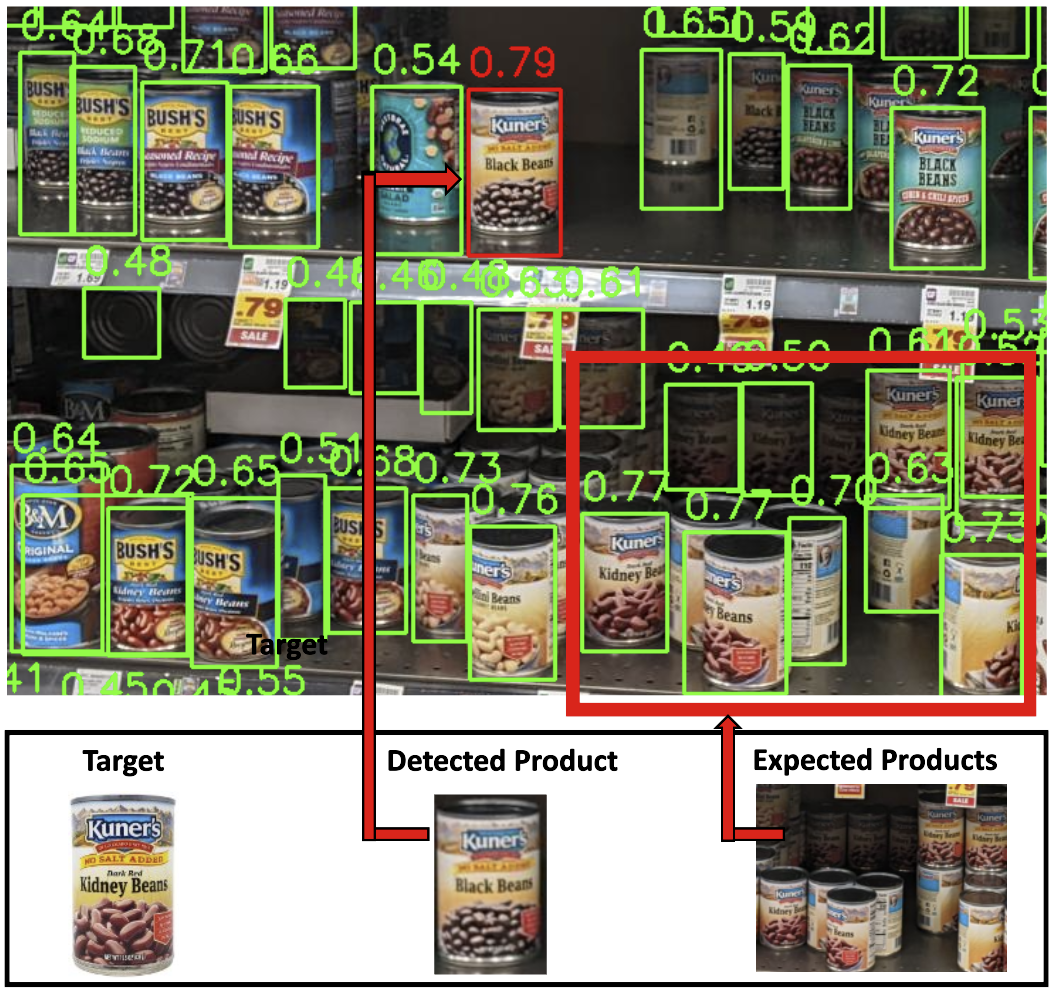}
    \caption{The product detection algorithm in a real grocery store. The system detects an incorrect product which is visually similar to the target. The expected products were either not facing straight or occluded partially by price tags hampering their visual similarity with the target.}
    \label{fig:discussion}
    \vspace{-4mm}
\end{figure}

Each participant retrieved 5 different products in each condition \textit{(continuous, discrete, human)}. Perhaps unsurprisingly, the human condition was successful \textbf{75/75} times. We observed a few scenarios where participants picked up the wrong product initially but were corrected by the human caller. 
The product locator system was the same for both the \textit{continuous} and the \textit{discrete} modes. The product detection failed \textbf{21/150} times in locating the desired product in the participant's first attempted scan of the shelf. This mostly happened when the desired product was not in the camera frame. If similar enough products are in frame (above threshold), the system selects the most visually similar product it can find. In some cases, when the desired target product is not facing straight or is occluded, the system might locate a visually similar product that may be incorrect (Figure \ref{fig:discussion}). While inconvenient, these errors are ultimately correctable using existing work (e.g., barcode based methods). The system showed promise in dealing with overexposure and blur due to the data association as shown in the example (Figure \ref{fig:blur}).  

Both of our planners guided the user to the desired product \textbf{150/150} (100\%) times. In the continuous mode, the participant picked up the adjacent item \textbf{8/75} times, and in the discrete mode that happened \textbf{6/75} times. It is important to note that the guidance algorithm took the participant close to the product, but since the users used their non-dominant hand to pick up the product they picked up a product immediately adjacent to the desired one (which was usually just 1-2 inches away). We include these data in our subsequent analysis, as the high level guidance was still successful in minimizing the exhaustive search required in the haptic space of the participant. 
\subsubsection{Hypotheses Testing}
\begin{figure}[h]
    \centering
    \includegraphics[width=180pt]{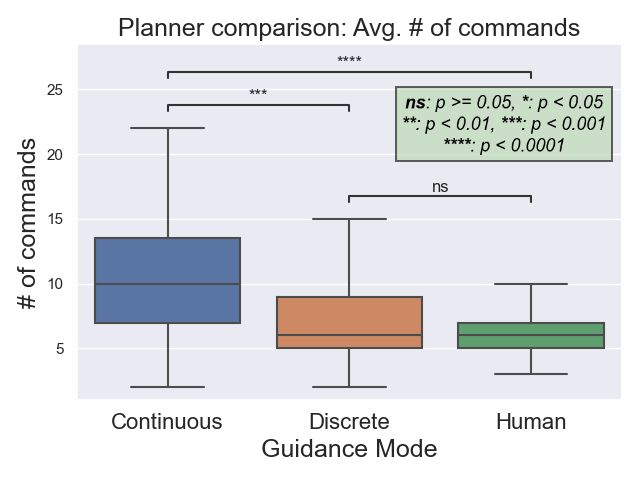}
    \caption{Average number of commands used by the different planners.}
    \label{fig:n_commands}
    \vspace{-4mm}
\end{figure}

\begin{figure}[h]
    \centering
    \includegraphics[width=160pt]{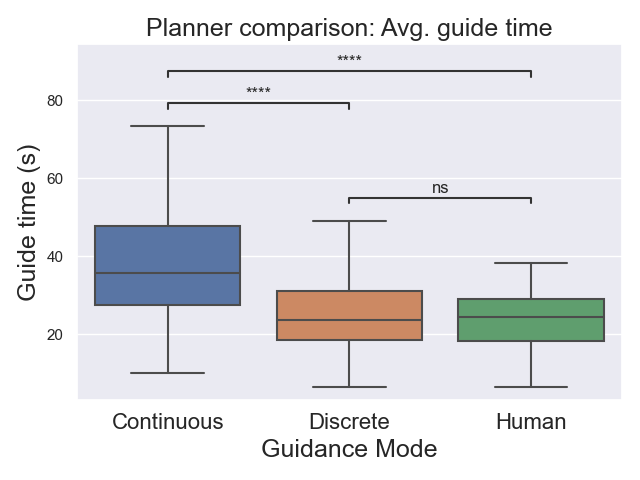}
    \caption{Average guide time used by the different planners.}
    \label{fig:guide_time}
    \vspace{-4mm}
\end{figure}

\begin{figure}[h]
    \centering
    \includegraphics[width=160pt]{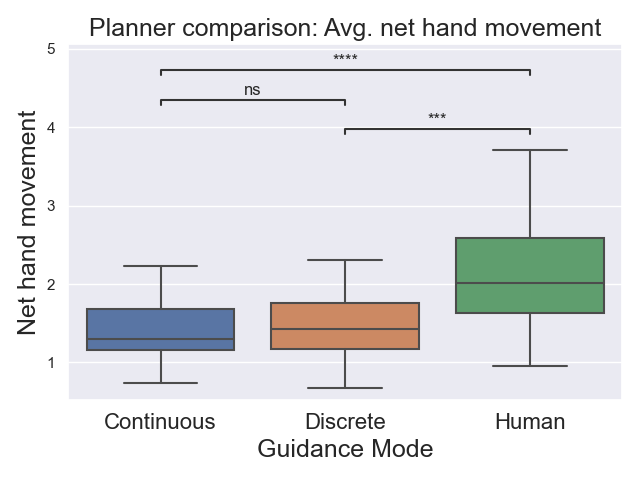}
    \caption{Average net hand movement caused by the different planners.}
    \label{fig:dist}
    \vspace{-4mm}
\end{figure}

\begin{figure}[h]
    \centering
    \includegraphics[width=160pt]{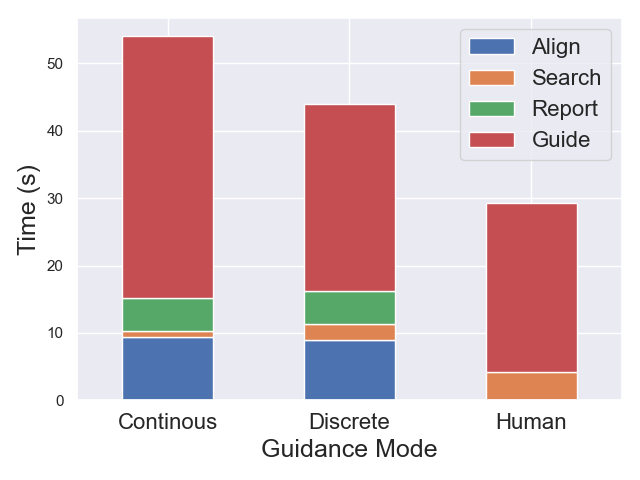}
    \caption{Average time taken by each planner. The alignment, search, and report durations are similar for Continuous and Discrete but the guidance time is lower for Discrete, comparable to human levels.}
    \label{fig:stacked}
    \vspace{-4mm}
\end{figure}

Post-hoc comparisons using Tukey's HSD test revealed that participants retrieved the products with significantly fewer commands in the discrete mode as compared to the continuous mode, \textbf{confirming H1} (Figure \ref{fig:n_commands}). In fact, the number of commands was similar for the human caller and the discrete mode. Using the same test we found that the participants could retrieve the products significantly faster in the discrete mode compared to the continuous mode, thus \textbf{confirming H2} (Figure \ref{fig:guide_time}). The guidance time for the discrete was also similar to the human caller. We ran a TOST (two one-sided test) which \emph{confirms distributional equivalence for the number of commands and guidance time between the discrete planner and human caller}. Although the same doesn't hold true if we consider the total time which includes: alignment, searching for the product, reporting the plan, and guiding. We can see (Figure \ref{fig:stacked}) that both proposed modes incur some time in aligning and reporting the plan. We did not classify all of these activities for the human caller because they would interleave these instructions into their guidance, so we classify the time prior to providing actual movement instructions as `search time'. We did not find any statistical difference between the net hand movement caused by either planner, thus we \textbf{could not confirm H3}. To our surprise, we see that the human caller had significantly higher net movement (Figure \ref{fig:dist}). We think that this is due to the fact that the human caller relied on tracking the dominant hand of the participant in order to give them instructions with respect to the dominant hand position. The dominant hand would often go out of the cell phone camera frame and the human caller would have to issue instructions in order to gauge where the hand is causing these superfluous movements. 

\subsubsection{Subjective Evaluation}
\begin{figure}[h]
    \centering
    \includegraphics[width=220pt]{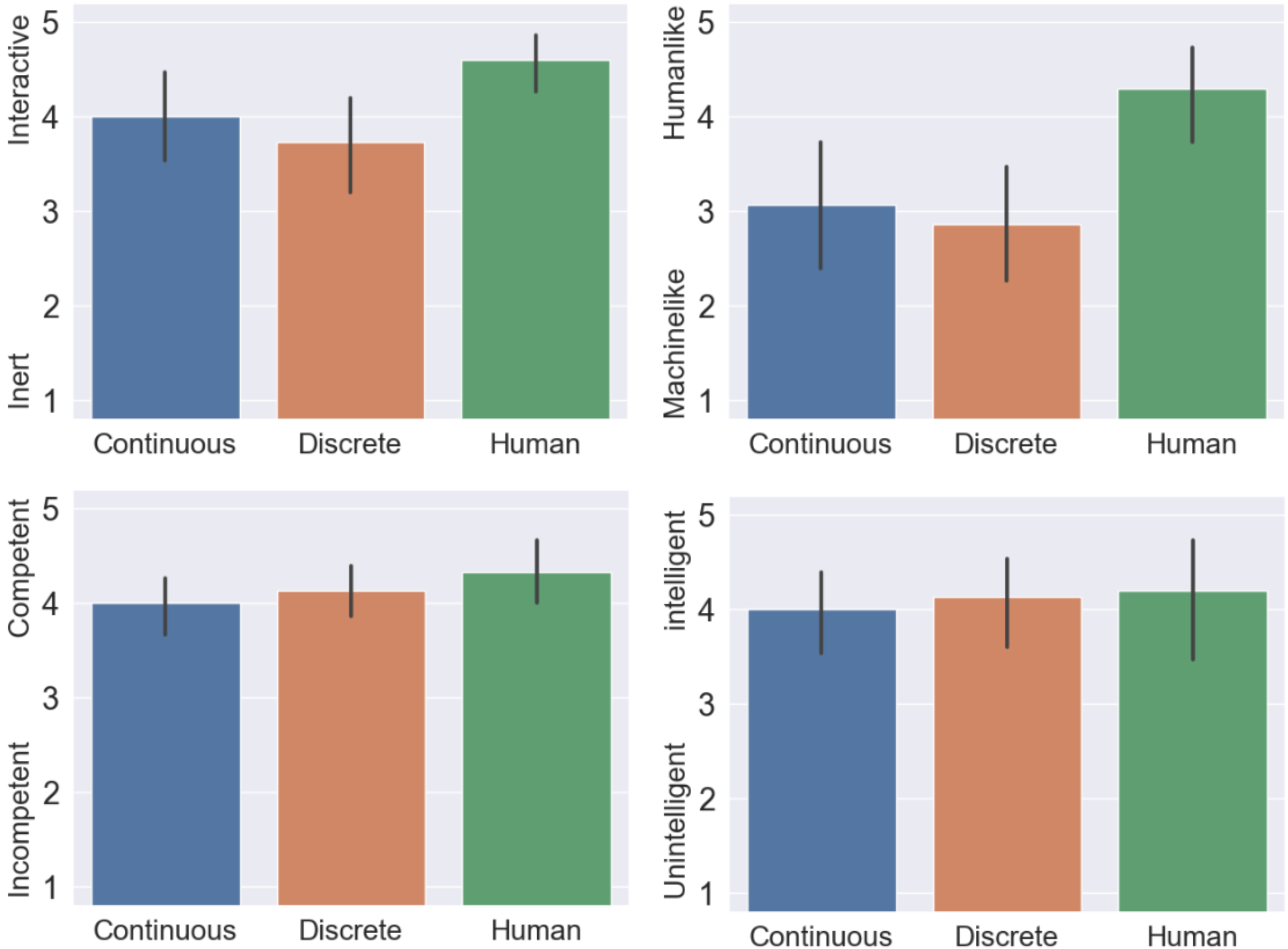}
    \caption{Average rating on subjective metrics: \textit{(clockwise from upper-left) Interactive, Humanlike, Intelligent, Competent}. Both planners performed well on all the metrics compared to the human guide except for \textit{Humanlike}.}
    \label{fig:sentiment}
    \vspace{-4mm}
\end{figure}

We administered a survey after each condition. Participants rated both of our planners high on metrics concerning: human-like, interactive, competent, and intelligent (Figure \ref{fig:sentiment}). The two planners and the human performance were not statistically different on the competence and intelligence metrics. Both of the planners were rated less humanlike than our human baseline. The discrete planner was rated slightly lower in the interactive metric which could be because the discrete planner did not provide any affirmations, which is an area of future investigation. The participants rated the plan overview's helpfulness as 4.2 (std 1.01) out of 5, indicating that this component was valuable to them. The large standard deviation is attributed to two participants rating this feature very low and in the exit interview they mentioned that it was hard and confusing for them to understand the overview template. We noticed participants mentioning a clear preference for one or the other planner. Some participants who liked the continuous mode mentioned that they liked that it provided some affirmation in the form of \textit{``Stop"} command. We also noticed the human caller using some kind of affirmation along with the movement instructions, for example, \textit{``Alright"}, \textit{``Good"}, \textit{``That one"}. Participants who liked the discrete planner more mentioned that they liked getting precise movement instructions and not relying on the system to stop them without much certainty about how much they would have to move.
Participants also rated ShelfHelp favorably on metrics such as \textit{ease of use, confidence, mental demand, temporal demand, frustration}.

\section{Discussion and Future Work}
We caution the reader to exercise care to avoid drawing strong conclusions about device readiness based solely on pilot tests with blindfolded, non-BVI participants, as these are not necessarily a reliable proxy for the (eventual) target population of this device \cite{Silverman-harmfuleffects,dosSantos-comparativeStudy}, but we find the preliminary results promising and indicative of value in further investigation.
With this pilot study, we have shown a proof-of-concept with regard to the hardware and software features of our proposed self-contained system (i.e., not dependent on external compute or data connection), alongside a validation of a novel product locator system coupled with a novel fine-grain manipulation guidance system, demonstrating that the system is able to locate a target product, plan to reach it, and effectively guide retrieval in real-time.
The discrete planner performs quantitatively better than the continuous and is on par with the human in our study. The continuous planner elicited a positive response as well, in part due to the affirmations it provides by using the \textit{``Stop"} command. It could be beneficial to add this feature to the discrete planner, effectively grounding the user during the execution of the plan. The product locator system relies on visual features which do not effectively leverage all of the available semantic information on product packaging, and a future direction would be real-time incorporation of semantic information as well. 

\section{Conclusion}
In this work, we present our system ShelfHelp which enhances an instrumented cane meant for navigational assistance with the addition of manipulation assistance capabilities. It includes a novel, maintenance-free product locator system that does not need regular re-training as new products are introduced, and works offline making it scalable and practical. We present a novel fine-grain manipulation guidance planner that effectively guides manipulation-based reachability in the haptic space of the user.
We tested aspects of the system using a pilot study with novice users that provided an initial validation for ShelfHelp and show that our discrete guidance planner optimizes for guide time and the total number of commands without compromising legibility. We tested two guidance systems with a quantitative and qualitative comparison against each other and a human baseline. The discrete guidance mode was on par with the human caller in terms of guide time, the number of commands, perceived competence, and intelligence but lacked in terms of interactiveness in form of affirmation. The study also surfaced positive feedback for qualitative metrics such as ease of use, confidence, mental demand, temporal demand, frustration, intelligence, and competence for the system with both of the guidance planners.






\bibliographystyle{ACM-Reference-Format} 


\end{document}